\pdfoutput=1

\documentclass[11pt]{article}

\usepackage{acl}

\usepackage{times}
\usepackage{latexsym}

\usepackage[T1]{fontenc}

\usepackage[utf8]{inputenc}

\usepackage{microtype}

\usepackage{inconsolata}

\usepackage{graphicx}
\usepackage{amsmath}
\usepackage{times}
\usepackage{framed}
\usepackage{ragged2e}
\usepackage{bm}
\usepackage{soul}
\usepackage{latexsym}
\usepackage{subfigure}
\usepackage{amsthm}
\usepackage{graphicx}
\usepackage{float}
\usepackage{longtable}
\usepackage{booktabs} 
\usepackage{multirow}
\usepackage{multicol}
\usepackage{amssymb}
\usepackage{dashbox}%
\usepackage{multirow}
\usepackage{textcomp}
\usepackage{tcolorbox}
\usepackage{bbding}
\usepackage{bbm}
\usepackage[ruled, vlined, linesnumbered]{algorithm2e}
\usepackage{mathtools}
\usepackage{adjustbox}
\usepackage[ruled,vlined]{algorithm2e}
\usepackage{color, soul}
\usepackage{pifont}
\usepackage{array}
\newcolumntype{P}[1]{>{\centering\arraybackslash}p{#1}}
\newcolumntype{M}[1]{>{\centering\arraybackslash}m{#1}}
\usepackage{cleveref}
\crefname{section}{§}{§§}
\Crefname{section}{§}{§§}
\crefname{figure}{Figure}{Figure}
\Crefname{figure}{Figure}{Figure}
\crefname{table}{Table}{Table}
\Crefname{table}{Table}{Table}
\usepackage{tabularx}
\newcommand\ourmethod{\textsc{RewardAgent}\xspace}
\newcommand\ourmethodmini{\textsc{RewardAgent}\textsubscript{\textsc{mini}}\xspace}
\newcommand\ourmethodllama{\textsc{RewardAgent}\textsubscript{\textsc{Llama}}\xspace}
\newcommand\ourdataset{\textsc{IFBench}\xspace}

\usepackage{colortbl}

\title{Agentic Reward Modeling: Integrating Human Preferences with \\ Verifiable Correctness Signals for Reliable Reward Systems}

\author{Hao Peng\thanks{\quad Equal contribution.}, Yunjia Qi$^{*}$, Xiaozhi Wang, Zijun Yao, Bin Xu, Lei Hou, Juanzi Li\\
Department of Computer Science and Technology, Tsinghua University \\
\texttt{\{peng-h24\}@mails.tsinghua.edu.cn}}

\begin{document}
\maketitle
\begin{abstract}

Reward models (RMs) are crucial for the training and inference-time scaling up of large language models (LLMs). 
However, existing reward models primarily focus on human preferences, neglecting verifiable correctness signals which have shown strong potential in training LLMs. In this paper, we propose \textit{agentic reward modeling},  a reward system that combines reward models with verifiable correctness signals from different aspects to provide reliable rewards. We empirically implement a reward agent, named \ourmethod, that combines human preference rewards with two verifiable signals: factuality and instruction following, to provide more reliable rewards. We conduct comprehensive experiments on existing reward model benchmarks and inference time best-of-n searches on real-world downstream tasks. \ourmethod significantly outperforms vanilla reward models, demonstrating its effectiveness.
We further construct training preference pairs using \ourmethod and train an LLM with the DPO objective, achieving superior performance on various NLP benchmarks compared to conventional reward models. Our codes are publicly released to facilitate further research\footnote{\url{https://github.com/THU-KEG/Agentic-Reward-Modeling}}.
\end{abstract}

\section{Introduction}

\begin{figure}[t]
    \centering
    \includegraphics[width=0.95\linewidth]{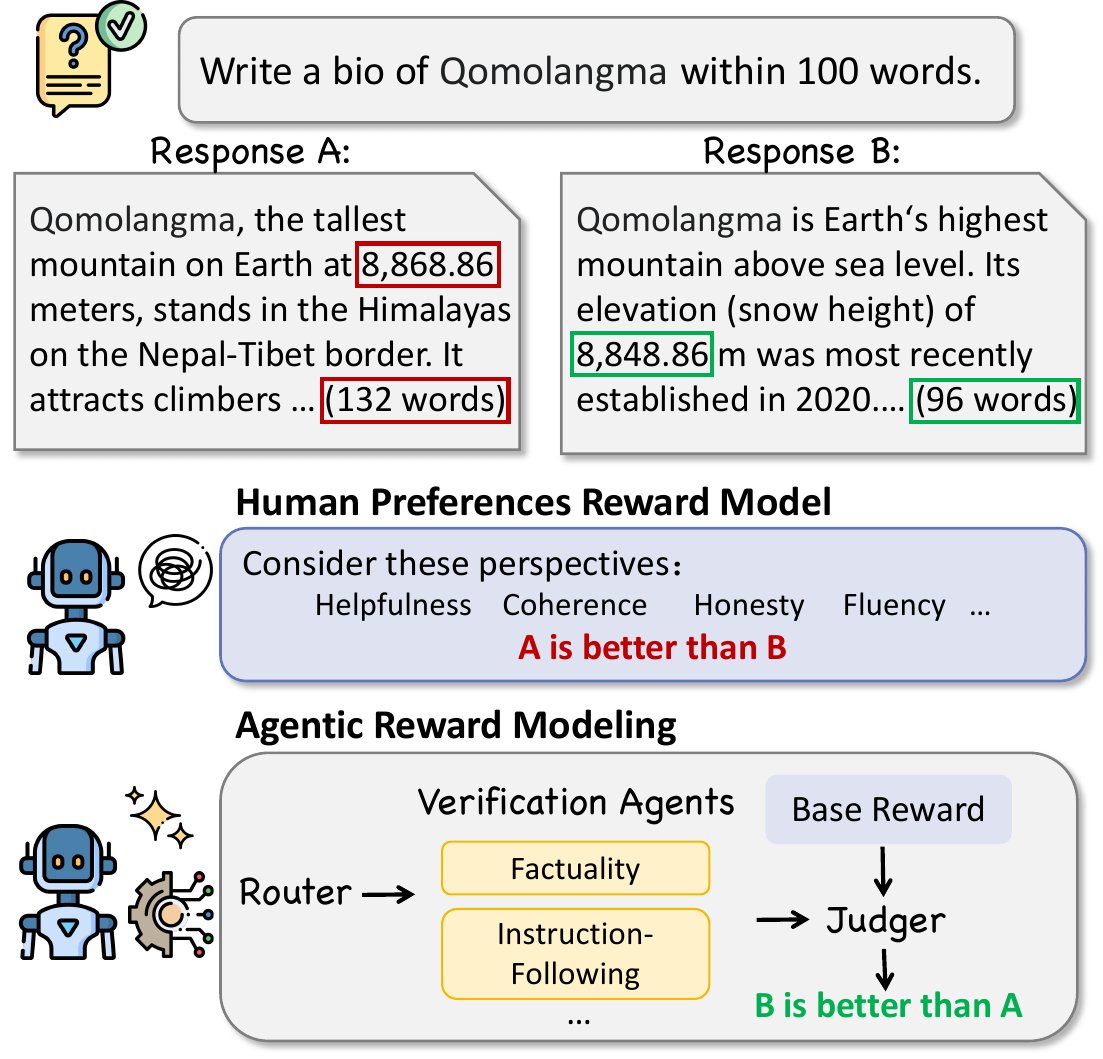} 
    \caption{An illustration of \textit{agentic reward modeling}.
    }
    \label{fig:fig1}
\end{figure}

Reward models (RMs) are designed to score the quality of responses and are typically used in the post-training of large language models (LLMs), such as RL~\citep{ouyang2022training} and DPO training~\citep{rafailov2024direct}, and in inference-time scaling laws~\citep{wu2024inference,snell2024scaling}, such as best-of-n search~\citep{brown2024large}. Reliable RMs are key to the success of modern LLMs.

Despite the success of reward models, existing RMs primarily focus on human preferences, which may be susceptible to subjective biases~\citep{saito2023verbosity,singhal2023long}, while neglecting verifiable correctness signals like factuality~\citep{liu2024rm, tan2024judgebench}. As illustrated in Figure~\ref{fig:fig1}, existing RMs may prefer the response \textit{A} due to its language style and longer length~\citep{singhal2023long}, overlooking factual errors and failure to follow instructions. This could affect the reliability of reward models and further influence the reliability of the trained LLMs~\citep{singhal2023long,chen2024odin}.
Conversely, verifiable correctness rewards exhibit notable potential in specific scenarios~\citep{guo2025deepseek}, providing a valuable complement to conventional reward models.

Based on the above considerations, we propose \textit{agentic reward modeling}, a reward system that combines reward models with verifiable correctness signals from different aspects to provide more reliable rewards. For example in Figure~\ref{fig:fig1}, a verification agent that specifically provides correctness signals, such as rule-based rewards~\citep{mu2024rule}, can be used to assess factual accuracy or verify adherence to instruction constraints. By integrating verifiable correctness rewards with human preferences, the reward system selects the superior response \textit{B}.
Agentic reward modeling enhances reliability through multi-dimensional correctness signals, enables flexible integration of diverse verification agents, and improves the interpretability of the final reward.

In this paper, we empirically implement a reward agent, named \ourmethod, which integrates the conventional human preference-based reward models with correctness signals from two key aspects: (1) factuality, which assesses the factual correctness of the claimed facts in the response, and (2) instruction-following, which evaluates whether the response adheres to the hard constraints in the instruction~\citep{zhou2023instruction}, such as length constraints, which significantly impact user experience in real-world applications~\citep{sun2024conifer, qi2024constraint}.
The architecture of \ourmethod is shown in Figure~\ref{fig:fig1}, consisting of three main modules: (1) \textit{Router}, which analyzes the instruction to determine the appropriate verification agents to invoke. (2) \textit{Verification agents}, which evaluate the correctness of response in different aspects, including factuality and instruction-following. Specifically, for factuality, we design a verification agent that efficiently evaluates factual correctness compared to the previous factuality evaluation framework~\citep{min2023factscore} through a process including pairwise comparison, query generation, evidence generation, and verification, where evidence generation 
can utilize either a search engine or the model's parametric knowledge. For instruction-following, we design a verification agent that extracts hard constraints, generates constraint checker code, and executes the code for verification, where the constraint checker is the Python code script to verify whether a given response satisfies a specific hard constraint.
(3) \textit{Judger}, which integrates the correctness signals from the verification agents and human preference scores from the reward models to provide an overall reward score.
We adopt ArmoRM~\citep{wang2024interpretable} as the reward model for computing human preference scores in \ourmethod. We use GPT-4o mini~\citep{OpenAI2024} and Llama3-8B Instruct~\citep{dubey2024llama} as the backbone LLMs for all the modules and implement \ourmethodmini and \ourmethodllama, respectively, except that in \ourmethodllama, the LLM backbone of the instruction-following agent is Qwen2.5-Coder 7B~\citep{hui2024qwen2}.

We conduct extensive experiments to validate the effectiveness of \ourmethod. First, we conduct an evaluation on several reward model benchmarks, including RM-Bench~\citep{liu2024rm} and JudgeBench~\citep{tan2024judgebench}, as they contain response pairs that involve factual correctness, and IFBench, which is newly constructed for instruction-following and contains $444$ instances, each of which includes an instruction with several hard constraints, a chosen response that satisfies all constraints, and a rejected response that violates some constraints. \ourmethod significantly outperforms other advanced reward models on these benchmarks. We further apply reward models in real-world downstream tasks, including inference-time best-of-n search and constructing training preference pairs.
We evaluate best-of-n search on factuality question answering dataset TriviaQA~\citep{joshi2017triviaqa} and instruction-following datasets, IFEval~\citep{zhou2023instruction} and CELLO~\citep{he2024can}. We adopt Llama3-8B Instruct and GPT-4o~\citep{OpenAI20244o} as policy models to generate $32$ responses for each instruction with $1.0$ sampling temperature. 
\ourmethod significantly outperforms the base reward model AromRM in the best-of-n search, suggesting its ability to select superior responses and unleash inference scaling laws.
Finally, we apply \ourmethod to construct training preference pairs and train an LLM using DPO~\citep{rafailov2024direct}.
Specifically, we construct training data from two sources: UltraFeedback~\citep{cui2024ultrafeedback} and on-policy data. We adopt Zephyr-7B~\citep{tunstall2023zephyr} as the policy model and train it using DPO. The LLM trained on \ourmethod-constructed data consistently outperforms those trained on AromRM annotations on several NLP benchmarks, which further demonstrates the effectiveness of \ourmethod. 
We encourage the community to explore more verifiable correctness signals to develop reliable reward systems for LLM development and alignment.
\section{Preliminaries}
In the LLM domain, a reward model is typically a regression model that takes an instruction and a response as input and outputs a reward score~\citep{ouyang2022training}, which can be formulated as $r_{\text{RM}}(x,y)$, where $x$ denotes an instruction and $y$ represents a response. Reward models are typically trained on a large set of preference pairs based on the Bradley-Terry (BT) model~\citep{bradley1952rank}.

However, due to the subjectivity and complexity of human preferences and the capacity limitations of the BT model~\citep{munos2023nash, swamy2024minimaximalist, sun2024rethinking}, reward models often exhibit subjective bias, such as favoring longer and detailed outputs~\citep{saito2023verbosity}, while neglecting verifiable correctness signals like factuality~\citep{liu2024rm, tan2024judgebench}. On the other hand, training LLMs with verifiable correctness signals has shown strong potential~\citep{lambert2024t, guo2025deepseek}. Based on these considerations, we propose \textit{agentic reward modeling}, a reward system that integrates reward models with verifiable correctness signals from different aspects to provide more reliable rewards. Agentic reward modeling can be formulated as follows:
\begin{equation}
r(x, y) = \underbrace{\lambda \cdot r_{\text{RM}}(x, y)}_{\text{base reward}} + \sum_{i \in A_x} \underbrace{w_i \cdot a_i(x, y)}_{\text{correctness signals}}
\label{eq:eq1}
\end{equation}
$\lambda$ denotes the weight of the base reward model. $a_i$ denotes a specific verification agent that provides verifiable correctness signals, such as rule-based rewards~\citep{mu2024rule}. $w_i$ denotes the corresponding weight for each verification agent, which can be set as a hyper-parameter or adaptive to the instruction. $A_x$ is an index subset of the complete set of verification agents $A$ and is determined based on the instruction $x$. Equation~\ref{eq:eq1} provides the fundamental concept of agentic reward modeling, which can be implemented in various ways to construct a reward agent and our implementation is in \cref{sec:method}.

\section{\ourmethod}
\label{sec:method}

\begin{figure*}[t]
    \centering
    \includegraphics[width=0.95\linewidth]{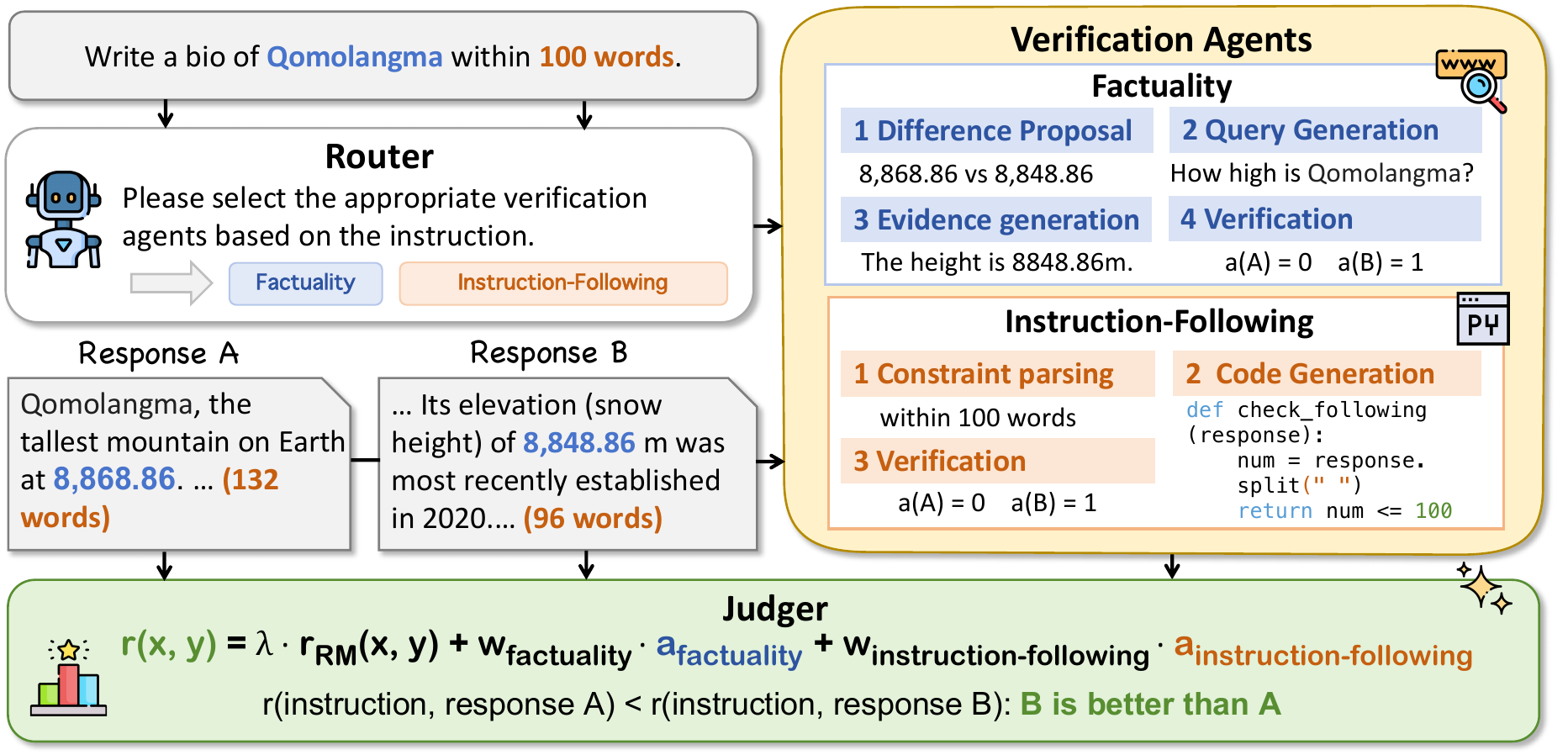} 
    \caption{The framework of \ourmethod, including three modules: Router, Verification Agents, and Judger.
    }
    \label{fig:pipeline}
\end{figure*}

In this work, we empirically implement a reward agent, named \ourmethod, which integrates the base human preference reward model with verifiable correctness signals from two key aspects: factuality, which assesses the correctness of claimed facts, and instruction-following, which evaluates whether the response satisfies the hard constraints specified in the instruction~\citep{zhou2023instruction}. 
Both aspects significantly impact reliability and user experience in real-world applications and are challenging to evaluate effectively with existing reward models~\citep{liu2024rm}.
This section introduces the overall model architecture (\cref{sec:model_archi}) and the specific modules (\Cref{sec:router,sec:scoring_agents,sec:judger}) of \ourmethod.

\subsection{Model Architecture}
\label{sec:model_archi}
Following the concept in Euqation~\ref{eq:eq1}, the overall architecture of \ourmethod is illustrated in Figure~\ref{fig:pipeline}, which consists of three main modules: (1) \textit{Router}, which analyzes the instruction and determines which agents to invoke, corresponding to $A_x$ in Equation~\ref{eq:eq1}. As different instructions may require evaluations of different aspects of responses, dynamically selecting verification agents helps reduce inference costs and mitigate potential cumulative errors.
(2) \textit{Verification agents}, which evaluate different aspects of response correctness. In our implementation, we design two agents for assessing factuality and instruction-following, both based on LLMs augmented with additional tools. 
(3) \textit{Judger}, which integrates the scores from the verification agents and human preferences from the base reward model to produce a final reward, corresponding to determining $\lambda$ and $w_i$ in Equation~\ref{eq:eq1}.
We will provide detailed descriptions in the following sections.

\subsection{Router}
\label{sec:router}
Given an instruction, the router analyzes its requirements to the response to select the appropriate verification agents. 
The router is powered by an existing LLM backbone. Specifically, we first manually provide a concise description for each verification agent, 
explaining its functionality and specifying the conditions for its usage. Then, we input the instruction with all agent descriptions into the LLM, prompting it to select appropriate verification agents for correctness assessment. More implementation details are placed in appendix~\ref{sec:app_method}.

\subsection{Verification Agents}
\label{sec:scoring_agents}
\paragraph{Factuality}
Previous studies have proposed various methods to evaluate the factuality of responses, such as FactScore~\citep{min2023factscore}, which can be directly used as a verification agent. However, these methods typically require extensive search engine queries to verify the correctness of each atomic fact, which is costly and inefficient for reward scoring. Intuitively, pairwise scoring based on only the differences between two responses can effectively reduce search engine queries and time costs. Therefore, we propose a pairwise factuality verification agent for efficiently evaluating the factual correctness of response pairs.
The agent is illustrated in Figure~\ref{fig:pipeline}, which consists of four main components:
(1) Difference proposal, which identifies key differences in claimed facts between two given responses.
(2) Query generation, which constructs queries based on the identified differences to retrieve evidence for distinguishing these differences.
(3) Evidence generation, which uses the generated queries to retrieve supporting evidence using either external search engines or parametric knowledge in LLMs.
(4) Verification, which assigns an integer score from $0$ to $1$ to each response, using the collected evidence and original responses as inputs.
The verification agent can effectively capture subtle factuality differences~\citep{jiang2023llm} between responses while significantly reducing inference-time costs by verifying only their differences rather than all claimed facts.
All modules are implemented using an LLM backbone. The implementation details are placed in appendix~\ref{sec:app_method}.


\paragraph{Instruction-Following}
The evaluation of the instruction following primarily assesses the adherence to hard constraints~\citep{zhou2023instruction} specified in the instruction, such as length constraints. 
Typically, instruction-following constraints can be categorized into soft and hard constraints, where the former focuses on semantic aspects, such as language style, while the latter focuses on surface-form constraints, such as format, which can be objectively evaluated. For instruction-following, our verification agent focuses on hard constraints, which are difficult to evaluate with existing reward models but can be efficiently verified using external tools, such as Python code scripts.
The agent is shown in Figure~\ref{fig:pipeline}, including three components:
(1) Constraint parsing, which extracts hard constraints from the instruction.
(2) Code generation and refinement, which generates Python scripts used to check the adherence to the extracted constraints. The generated code takes the response as input and returns either $0$ or $1$, where $1$ indicates that the constraint is satisfied, and $0$ otherwise.
We also incorporate a refinement step like \citet{madaan2024self} to correct invalid or syntactically incorrect code. Specifically, we execute the generated Python code using a Python interpreter, and if an error occurs, the error information and original code are fed back into the model to generate a refined code script.
(3) Verification, which executes the generated code in the Python interpreter to obtain a binary score ($0$ or $1$). The final score is the average of all hard constraint scores.
All the modules are implemented using LLMs. The specific prompts and implementation details are provided in appendix~\ref{sec:app_method}.

\subsection{Judger}
\label{sec:judger}
The judger integrates reward scores from verification agents and human preferences from base reward models. In our implementation, we use a weighted sum as the judger, where $\lambda$ and $w_i$ are all set to $1.0$, to compute the final reward score in Equation~\ref{eq:eq1}. One can also adopt different $\lambda$ and $w_i$ for better applicability in different scenarios. Additionally, the judger can dynamically adjust $\lambda$ and $w_i$ based on the instruction like gating network~\citep{wang2024interpretable}, we leave it as future work.

\section{Experiments}
This section presents experiments on several reward model benchmarks, including experimental setup (\cref{sec:exp_setup}), results (\cref{sec:exp_result}), and analyses (\cref{sec:exp_analysis}).

\begin{table*}
    \centering
    \small
    \begin{tabular}{lccccccc}
    \toprule
    \multirow{2}{*}{Model} & \multicolumn{2}{c}{RM-Bench} & \multirow{2}{*}{JudgeBench} & \multicolumn{3}{c}{IFBench} & \multirow{2}{*}{Overall} \\
    \cmidrule{2-3} \cmidrule{5-7}
    & Normal & Hard & & Simple & Normal & Hard & \\
    \midrule
ArmoRM-Llama3-8B-v0.1 &$76.7$&$34.6$&$51.9$&$72.3$&$66.2$&$59.5$&$56.5$\\
INF-ORM-Llama3.1-70B &$77.5$&$25.1$&$59.1$&$78.7$&$69.2$&$53.8$&$55.7$\\
Skywork-Reward-Llama-3.1-8B-v0.2 &$78.0$&$31.8$&$57.8$&$78.7$&$69.2$&$59.8$&$58.1$\\
Skywork-Reward-Gemma-2-27B &$82.7$&$35.1$&$55.8$&$\boldsymbol{87.2}$&$68.4$&$56.1$&$59.2$\\
internlm2-7b-reward &$72.6$&$19.9$&$56.2$&$74.5$&$61.7$&$55.7$&$52.0$\\
internlm2-20b-reward &$74.4$&$26.1$&$61.7$&$74.5$&$68.4$&$58.7$&$56.4$\\
\midrule
GPT-4o &$71.4$&$27.9$&$64.6$&$\underline{85.1}$&$66.2$&$54.4$&$56.3$\\
GPT-4o mini &$60.5$&$15.0$&$51.9$&$70.2$&$59.4$&$51.9$&$45.9$\\
o3-mini &$76.0$&$38.6$&$66.6$&$81.9$&$\underline{76.3}$&$64.6$&$62.8$\\
Llama3-8B Instruct &$\phantom{0}9.3$&$20.2$&$\phantom{0}2.6$&$12.8$&$12.8$&$13.6$&$11.3$\\
DeepSeek-R1 &$83.7$&$50.1$&$\boldsymbol{74.4}$&$72.3$&$74.4$&$64.0$&$69.1$\\
DeepSeek-R1-Distill-Llama-8B &$42.1$&$56.8$&$47.7$&$53.2$&$55.6$&$54.2$&$50.3$\\
\midrule
\ourmethodllama &$79.3$&$53.5$&$52.9$&$70.2$&$63.9$&$67.8$&$63.2$\\
\quad w/ search engine &$76.0$&$49.9$&$55.2$&$74.5$&$69.2$&$67.8$&$62.5$\\
\ourmethodmini &$\boldsymbol{86.0}$&$\boldsymbol{60.2}$&$\underline{68.2}$&$78.7$&$69.2$&$\boldsymbol{78.0}$&$\boldsymbol{72.5}$\\
\quad w/ search engine &$\underline{84.2}$&$\underline{59.7}$&$60.7$&$68.1$&$\boldsymbol{80.5}$&$\underline{76.1}$&$\underline{70.3}$\\
    \bottomrule
    \end{tabular}
    \caption{Experimental results (\%) of all investigated baselines and \ourmethod. The overall score is the average of RM-Bench, JudgeBench, and the micro-averaged score of three subsets of IFBench. By default, \ourmethod relies on its parametric knowledge, and ``w/ search engine'' denotes using Google API as an external source.}
    \label{tab:main_exp}
\end{table*}

\subsection{Experimental Setup}
\label{sec:exp_setup}

\paragraph{\ourmethod Implementation}
We adopt the advanced and lightweight ArmoRM~\citep{wang2024interpretable} as the base reward model to compute human preference scores. As \ourmethod is agnostic to reward models, one can also adopt other advanced reward models. 
We use GPT-4o mini~\citep{OpenAI2024} as the LLM backbone for implementing all modules and developing \ourmethodmini. We also employ the open-source LLM Llama3-8B Instruct~\citep{dubey2024llama} as the backbone and develop \ourmethodllama, except for the instruction-following verification agent, which requires strong coding capabilities and is instead powered by Qwen2.5-Coder 7B~\citep{hui2024qwen2}.
We adopt two knowledge sources for the factuality verification agent: an external search engine using Google API and the LLM’s parametric parameters. More details are placed in appendix~\ref{sec:app_method}.

\paragraph{Evaluation Benchmarks}
Reward model benchmarks typically involve an instruction and a response pair and require selecting the better response as the chosen one.
We use RM-Bench~\citep{liu2024rm}, JudgeBench~\citep{tan2024judgebench}, and a new benchmark \ourdataset as evaluation benchmarks, as both RM-Bench and JudgeBench include response pairs involving factual correctness. We select the chat subset of RM-Bench as the evaluation set, using both the normal and hard settings. For JudgeBench, we use the knowledge subset as the evaluation set.
We further construct a new benchmark \ourdataset to evaluate reward models on selecting responses that better follow constraints in instructions as there is no existing relevant benchmark.
Specifically, we first construct instructions with several implicit constraints, integrating the constraint information with the primary task objective through paraphrasing. The constraints include both hard constraints, such as length, format, and keywords, as well as soft constraints, such as content and style. We then use GPT-4o to generate $8$ responses for each instruction with a sampling temperature of $1.0$. For each instruction, we create a response pair, selecting the one that satisfies all constraints as the chosen response and otherwise rejected. Based on the number of unsatisfied constraints (UC) in the rejected response, we split \ourdataset instances into three subsets: simple (\#UC$\geq$3), normal (\#UC$=$2), and hard (\#UC$=$1),  containing $47$, $133$, and $264$ instances respectively. 
We report the micro-averaged accuracy across the three subsets as the final metric for \ourdataset. More evaluation details on these benchmarks are provided in appendix~\ref{sec:app_exp}.

\paragraph{Baselines}
\looseness = -1
We mainly investigate two categories of baselines: (1) typical reward models, which are specifically trained for reward modeling and typically implemented as regression models to score each response and select the one with the highest reward score as the chosen response. We investigate several advanced and representative reward models, including ArmoRM~\citep{wang2024interpretable}, INF-ORM-Llama3.1-70B~\citep{infly2024inf}, Skywork-Reward~\citep{liu2024skywork}, internlm2 reward~\citep{cai2024internlm2}. (2) LLMs as generative reward models, where large language models serve as generative reward models to score responses or perform pairwise comparisons to select the best response~\citep{lambert2024rewardbench}. We evaluate proprietary models, including GPT-4o~\citep{OpenAI20244o}, GPT-4o mini~\citep{OpenAI2024}, o3-mini~\citep{openai2025o3mini}, and open-source LLMs, including Llama3-8B Instruct~\citep{dubey2024llama}, 
DeepSeek-R1, and R1 distilled Llama3-8B model~\citep{guo2025deepseek}. 
We evaluate all the baselines using the code repository provided by \citet{lambert2024rewardbench}.

\subsection{Experimental Results}
\label{sec:exp_result}

Table~\ref{tab:main_exp} presents the experimental results, and we can observe that:
(1) Existing reward models fall short in selecting more factual responses or better adhering to hard constraints in instructions, which may limit their reliability in real-world applications.
(2) \ourmethod significantly outperforms the base reward model AromRM and the corresponding LLM backbone GPT-4o mini and Llama3-8B Instruct. It demonstrates that designing an appropriate reward agentic workflow can effectively enhance reward model performance.
(3) Even when using Llama3-8B Instruct as the LLM backbone, \ourmethodllama outperforms reward models with much more parameters and more advanced proprietary LLMs such as GPT-4o, which suggests that \ourmethod is more cost-efficient without requiring additional reward modeling training data or more parameters to achieve advanced performance.
(4) Using a search engine as an external knowledge source for factuality slightly reduces performance in RM-Bench and JudgeBench. One possible reason is that the retrieved information may contain noise or irrelevant information~\citep{chen2024benchmarking}. We leave the detailed analysis and design of retrieval-augmented agents for future work.
(5) \ourmethod achieves significant improvements on IFBench, particularly in the hard subset. It suggests that while not perfectly solved, existing LLMs can effectively analyze hard constraints and generate verification code, which can help the training of advanced LLMs~\citep{lambert2024t}.

In conclusion, incorporating additional verification agents for specific scenarios~\cite{mu2024rule, lambert2024t}, particularly those with verifiable correctness, can develop more reliable and advanced reward systems, presenting a promising direction for future reward model development.

\subsection{Analysis}
\label{sec:exp_analysis}

\begin{table}
    \centering
    \small
    \resizebox{\linewidth}{!}{
    \setlength{\tabcolsep}{3pt}
    \begin{tabular}{lccc}
    \toprule
    Model & RM-Bench & JudgeBench & IFBench \\
    \midrule
    \ourmethodmini &$73.1$&$68.2$&$75.5$\\
    \hspace{2mm}\textit{-- factuality verifier} &$54.0$&$52.9$&$73.6$\\
    \hspace{2mm}\textit{-- if verifier} &$74.7$&$66.2$&$60.4$\\
    \hspace{2mm}\textit{-- both} &$55.4$&$58.8$&$58.8$\\
    \midrule[0.1pt]
    Oracle setting &$76.7$&$70.1$&$77.5$\\
    \midrule
    \ourmethodllama &$66.4$&$52.9$&$66.9$\\
    \hspace{2mm}\textit{-- factuality verifier} &$51.9$&$51.6$&$65.8$\\
    \hspace{2mm}\textit{-- if verifier} &$58.0$&$57.5$&$57.2$\\
    \hspace{2mm}\textit{-- both} &$44.8$&$55.5$&$57.2$\\
    \midrule[0.1pt]
    Oracle setting &$79.5$&$73.1$&$68.5$\\
    \bottomrule
    \end{tabular}
    }
    \caption{Experimental results (\%) of ablation study and the oracle setting. \textit{-- factuality verifier} and \textit{-- if verifier} refer to the reduction of the corresponding verification agent into a single LLM scorer.
    The results are the micro-averaged scores of all the corresponding subsets.}
    \label{tab:analysis}
\end{table}

We first conduct an ablation study on the verification agents in \ourmethod. Specifically, we investigate three settings: \textit{-- factuality verifier}, \textit{-- if verifier}, and \textit{-- both}, where the corresponding verification agents are reduced to \textbf{a single step}: using an additional LLM backbone to directly score the response, which is equivalent to the simple ensemble of the reward model ArmoRM with the corresponding LLM as a generative reward model~\citep{costereward2024}.
The ablation results are shown in Table~\ref{tab:analysis}. We can observe that removing the well-designed verification agent leads to a significant performance decrease. It demonstrates the importance of well-designed verification agents, and we encourage the community to develop more advanced verification agents for a more reliable \ourmethod.

We also observe the oracle setting of \ourmethod that invokes the most appropriate verification agents, that is, invoking the factuality agent on RM-Bench and JudgeBench, and the instruction-following verification agent on IFBench. The experimental results are shown in Table~\ref{tab:analysis}, and we observe that both \ourmethodmini and \ourmethodllama perform significantly better in the oracle setting. This further demonstrates the effectiveness of the verification agents and suggests that the planner in \ourmethod still has a large room for improvement and we leave developing a more advanced planner for future work. This also suggests that in some specific and well-defined scenarios, one can adopt the corresponding verification agent alone to achieve better results.

\begin{figure*}
    \centering
    \includegraphics[width=0.98\linewidth]{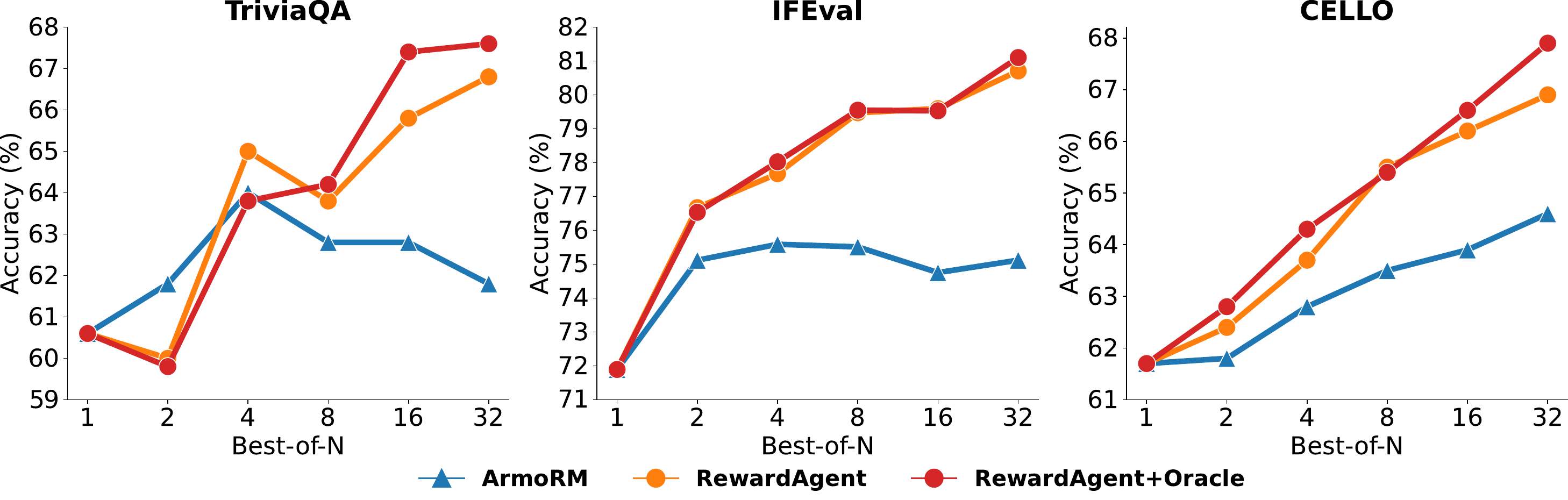}
    \caption{Best-of-n results (\%) on TriviaQA, IFEval, and CELLO using the base reward model ArmoRM and \ourmethod to search. ``+Oracle'' denotes using the oracle setting of \ourmethod as mentioned in \cref{sec:exp_analysis}.}
    \label{fig:enter-label}
\end{figure*}

\begin{table*}
    \centering
    \small
    \begin{tabular}{lccccccc}
    \toprule
    DPO Training Data & MMLU & MMLU-Pro & TriviaQA & TruthfulQA & IFEval & CELLO & MT-Bench\\
    \midrule
    -- & $58.9$ & $28.8$ & $54.8$ & $39.5$ & $43.3$ & $51.5$ & $5.2$ \\
    \midrule
    Original UF & $58.7$ & $29.3$ & $54.0$ & $42.0$ & $56.8$ & $62.0$  & $6.0$ \\
    ArmoRM-UF & $58.1$ & $29.9$ & $52.5$ & $45.0$ & $58.6$ & $60.8$ & $6.0$ \\
    \ourmethodllama-UF & $59.1$ & $30.5$ & $55.1$ & $44.1$ & $\mathbf{59.4}$ & $60.1$ & $5.8$ \\
    \midrule
    ArmoRM-OP & $58.4$ & $30.4$ & $51.6$ & $44.4$ & $52.7$ & $58.1$ & $6.0$ \\
    \ourmethodllama-OP & $\mathbf{59.5}$ & $\mathbf{31.3}$ & $\mathbf{55.3}$ & $\mathbf{48.5}$ & $58.2$ & $\mathbf{65.7}$ & $\mathbf{6.1}$ \\
    \bottomrule
    \end{tabular}
    \caption{Experimental results (\%) of LLMs trained with DPO on different training data. ``ArmoRM-UF'' denotes using ArmoRM to construct preference pairs from UltraFeedback. ``UF'' and ``OP'' are short for UltraFeedback and on-policy data, respectively. ``Original UF'' refers to using the original GPT-4 annotated preference pairs from UltraFeedback to train the LLM. ``--'' denotes the original LLM zephyr-7b-sft-full without further DPO training.}
    \label{tab:dpo_results}
\end{table*}

\section{Applications}
This section explores applying \ourmethod to inference-time search (\cref{sec:best_of_n}) and the training of LLMs (\cref{sec:dpo_train}) to further validate its effectiveness.

\subsection{Best-of-N Search}
\label{sec:best_of_n}

One important application of reward models is to conduct the inference-time search to find a better response~\citep{brown2024large,zhang2024generative}, which unleashes the inference-time scaling laws of LLMs~\citep{snell2024scaling, wu2024inference}. 
Therefore, we explore applying \ourmethod to the best-of-n search on downstream tasks. Specifically, we evaluate the best-of-n performance searched by \ourmethod on factuality question answering and constrained instruction following tasks.

\paragraph{Experimental Setup}
We conduct the best-of-n experiments on the factuality question answering dataset TriviaQA~\citep{joshi2017triviaqa}, and the instruction-following datasets IFEval~\citep{zhou2023instruction} and CELLO~\citep{he2024can}. We use Llama3-8B Instruct and GPT-4o as the policy models to generate $32$ responses for each instruction with $1.0$ sampling temperature. We perform best-of-n search using the base reward model ArmoRM~\citep{wang2024interpretable}, \ourmethodmini, and the oracle setting of \ourmethodmini.  The oracle setting refers to invoking the factuality verification agent on TriviaQA, and the instruction-following verification agent on IFEval and CELLO.

\paragraph{Experimental Results}
The results of the best-of-n experiments using Llama3-8B Instruct as the policy model are shown in Figure~\ref{fig:enter-label}. We can observe that \ourmethod significantly improves the best-of-n performance compared to using the base reward model ArmoRM, and the oracle setting further improves the results. 
It further validates the effectiveness of \ourmethod. 
The results using GPT-4o as the policy model are provided in appendix~\ref{sec:app_exp}, demonstrating the same trends and conclusions. We encourage the community to design more verification agents to unleash the inference scaling laws of LLMs across different scenarios.

\subsection{DPO Training}
\label{sec:dpo_train}
Reward models are primarily used to train LLMs using RL~\citep{ouyang2022training} or DPO~\citep{rafailov2024direct}. 
Considering RL training is resource-intensive, we explore employing \ourmethod to construct preference pairs for DPO training to validate its effectiveness in real-world applications.

\paragraph{Experimental Setup}
We construct two training datasets based on: (1) UltraFeedback~\citep{cui2024ultrafeedback}, where each instruction contains $4$ responses sampled from various LLMs. (2) on-policy, which contains $20,000$ instructions sampled from UltraFeedback and
each instruction contains $8$ responses sampled from the policy model itself with $1.0$ sampling temperature. We use reward models to score each response, taking the highest-scored response as the chosen one and the lowest as the rejected one to construct training pairs.
We adopt the zephyr-7b-sft-full~\citep{tunstall2023zephyr} model as the policy model to conduct DPO training because it is trained only using SFT~\citep{ouyang2022training}. We evaluate the DPO-trained LLMs on various NLP benchmarks, including MMLU~\citep{hendrycksmeasuring}, MMLU-Pro~\citep{wang2024mmlu}, TriviaQA~\citep{joshi2017triviaqa}, TruthfulQA~\citep{lin2022truthfulqa}, IFEval~\citep{zhou2023instruction}, CELLO~\citep{he2024can}, and MT-Bench~\citep{zheng2023judging}. More experimental details are provided in appendix~\ref{sec:app_exp}.

\paragraph{Experimental Results}

The experimental results are shown in Table~\ref{tab:dpo_results}. We can observe that LLMs trained with data constructed by \ourmethod generally outperform those trained with ArmoRM, especially on the factuality question answering and instruction-following datasets. The improvement is more significant in on-policy data.
Furthermore, models trained with \ourmethod-annotated data consistently outperform those trained on original UltraFeedback that is constructed with GPT-4. Notably, \ourmethodllama uses open-source Llama3-8B Instruct and Qwen2.5-Coder 7B as the LLM backbones, at a much lower cost than GPT-4.
The results further validate the effectiveness and applicability of \ourmethod. We believe using a more powerful LLM backbone in \ourmethod can achieve more advanced results and encourage the community to explore more advanced reward agents for better performance and reliability.

\section{Related Work}
Reward models are typically employed to score responses and are crucial to the success of modern LLMs. Since the emergence of RLHF~\citep{ouyang2022training}, numerous studies have focused on developing more advanced reward models to help train LLMs. The approaches mainly include designing model architectures~\citep{wang2024interpretable,dorka2024quantile,chen2025LDLRewardGemma} and utilizing more high-quality data or new training objectives~\citep{infly2024inf,yuan2024advancing,park2024offsetbias,liu2024skywork,cai2024internlm2,cao2024compass,lou2024uncertainty,litool2024,wang2024helpsteer2}. There are also various studies exploring using LLMs as generative reward models~\citep{zheng2023judging,mahan2024generative,skyworkcritic2024,cao2024compass,tan2024judgebench,yu2024self,alexandru2025atlaseleneminigeneral}.
Reward models are typically used for inference-time scaling laws~\citep{irvine2023rewarding,wu2024inference,snell2024scaling,brown2024large,xin2024deepseek} or for training, such as RL\citep{ouyang2022training} or DPO~\citep{rafailov2024direct}.

Despite the success of reward models, they primarily focus on human preferences, which may be susceptible to subjective biases or reward hacking~\citep{saito2023verbosity,singhal2023long,gao2023scaling,zhang2024lists,chen2024odin}. A notable limitation is \textit{verbosity bias}~\citep{saito2023verbosity}, where reward models tend to favor longer responses~\citep{singhal2023long, liu2024rm}. Additionally, some studies have shown that reward models may overlook correctness signals, such as factuality~\citep{lin2024flame, liu2024rm, tan2024judgebench}. These limitations affect the reliability of reward models, thereby impacting the performance of the trained LLMs~\citep{singhal2023long}.

Recently, several studies have shown that rule-based reward models or verifiable reward signals achieve impressive results in specific domains such as math~\citep{guo2025deepseek}, safety~\citep{mu2024rule}, instruction-following~\citep{lambert2024t}, medical~\citep{chen2024huatuogpt}, and finance~\citep{qian2025fino1}. The simplicity and advanced performance of rule-based reward models demonstrate significant potential for training LLMs, but it is still non-trivial to generalize to general domains.
In this paper, we explore combining human preferences from reward models with verifiable correctness signals to develop more reliable reward systems.
We believe that combining human preferences with verifiable correctness signals is a promising direction and encourage further research efforts in this area.

\section{Conclusion}

In this paper, we propose \textit{agentic reward modeling}, a reward system that integrates the human preferences from conventional reward models with verifiable correctness signals to provide more reliable rewards. We empirically implement a reward agent, named \ourmethod, which consists of a router, well-designed verification agents for factuality and instruction-following, and a judger. We conduct extensive experiments on reward modeling benchmarks, best-of-n search, and DPO training. \ourmethod significantly outperforms other reward models and LLMs as generative reward models. We encourage more research efforts to develop more advanced and reliable reward systems.
\section*{Limitations}

The main limitations of this work lie in the implementation of \ourmethod: (1) The verification agents are far from providing perfect rewards, as the average score on reward modeling benchmarks only reaches $72.5\%$. This suggests that achieving perfect rewards is challenging and requires further research efforts. (2) We only implement verification agents for factuality and instruction-following, which we believe are current weaknesses in reward models~\citep{liu2024rm} and important factors affecting LLM applications and user experiences. We encourage the community to explore more verifiable correctness signals. In conclusion, we believe the contribution of \textit{agentic reward modeling} concept is substantial, and we look forward to developing more advanced reward systems in the future.

\section*{Ethical Considerations}
We discuss the ethical considerations here:
(1) Intellectual property. 
We have strictly adhered to the licenses of all utilized artifacts, including datasets, models, and code repositories. We will open-source \ourmethod, code, and IFBench under the MIT license\footnote{\url{https://opensource.org/license/mit}}.
(2) Intended use and potential risk control.
We propose \textit{agentic reward modeling}, a reward system that integrates human preferences with correctness signals. We implement a reward agent named \ourmethod to provide more reliable rewards. We believe that all data used is well anonymized. Our model does not introduce additional ethical concerns but may provide incorrect rewards due to performance limitations. Users should not conduct reward hacking~\citep{skalse2022defining} and should carefully check important information.
(3) AI assistance. 
We have used ChatGPT to refine some sentences.

\bibliography{custom}

\newpage
\clearpage
\appendix
\section*{Appendices}
\section{\ourmethod Details}
\label{sec:app_method}

\looseness=-1
Tables~\ref{tab:planner} to~\ref{tab:if_agent} present the LLM prompts used for the implementation of \ourmethod. We employed Serper\footnote{\url{https://serper.dev/}} to implement our external search engine and we utilize the \texttt{gpt-4o-mini-2024-07-18} model in the \ourmethodmini version.
\section{Experimental Details}
\label{sec:app_exp}
In this section, we provide a detailed description of the evaluation process, divided into three parts: the construction and distribution details of \ourdataset~\ref{sec:app_exp_ifbench}, the evaluation dataset settings~\ref{sec:app_exp_evaluation}, and additional experimental results~\ref{sec:app_exp_more_res}.

\subsection{\ourdataset Details}
\label{sec:app_exp_ifbench}

\ourdataset is a benchmark designed to evaluate reward models for multi-constraint instruction-following. The dataset comprises $444$ carefully curated instances, each containing: an instruction with $3$ to $5$ multi-constraints, a chosen response satisfying all constraints, and a rejected response violating specific constraints. All instances were constructed using \texttt{gpt-4o-2024-11-20} version through the following systematic pipeline.

\looseness=-1
\paragraph{Instruction Construction} We sampled $500$ initial instructions from the Open Assistant~\cite{kopf2023openassistant}. To ensure clarity and simplicity, we constrained the initial instruction length to $5$ to $20$ words. Subsequently, we employed GPT-4o to generate five distinct categories of constraints for each initial instruction. It then autonomously selected $3$ to $5$ constraints and paraphrased them into $1$ to $2$ sentences. The paraphrased constraints were integrated into the initial instruction. Finally, we use GPT-4o to evaluate the final instructions and filter out those with internal contradictions, resulting in a final set of $444$ instructions.

\looseness=-1
\begin{itemize}
    \item {\bf Content Constraints: } Specify conditions governing response, including topic focus, detail depth, and content scope limitations.
    \item {\bf Style Constraints: } Control linguistic characteristics such as tone, sentiment polarity, empathetic expression, and humor.
    \item {\bf Length Constraints: } Dictate structural requirements including word counts, paragraph composition, and specific opening phrases.
    \item {\bf Keyword Constraints: } Enforce lexical constraints through keyword inclusion, prohibited terms, or character-level specifications.
    \item {\bf Format Constraints: } Define presentation standards that include specific formats such as JSON, Markdown, or Python, along with section organization and punctuation rules.
\end{itemize}

\begin{figure}[!ht]
    \centering
    \subfigure[]{
    \includegraphics[width=0.45\linewidth]{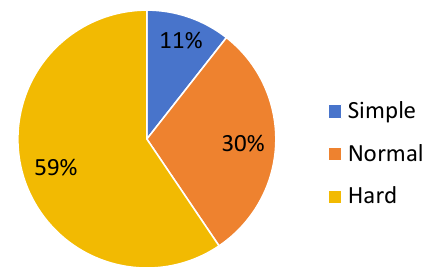} }
    \subfigure[]{
    \includegraphics[width=0.45\linewidth]{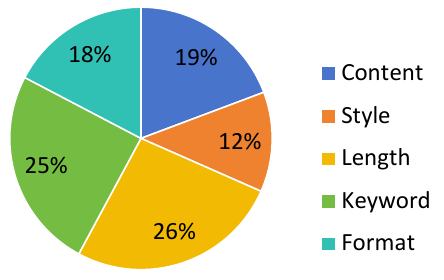} }
    \caption{Proportion (\%) of data in \ourdataset based on the number of unsatisfied constraints per instance and the categories of all unsatisfied constraints. }
    \label{fig:IFbench}
\end{figure}

\paragraph{Response Construction} For each instruction, we generated $8$ candidate responses using GPT-4o with temperature $1.0$ to maximize diversity. The chosen response was selected as the unique candidate satisfying all constraints through automated verification. Rejected responses were systematically selected to ensure balanced distributions of unsatisfied constraint (UC) categories and counts. As shown in Figure~\ref{fig:IFbench}, instances are stratified by difficulty: simple (\#UC$\geq$3), normal (\#UC$=$2), and hard (\#UC$=$1), with detailed information of UC category distributions. Specifically, (a) shows the distribution by the number of unsatisfied constraints in the rejected responses, where the sum of all parts equals the total number of instances. (b) presents the distribution by the categories of all unsatisfied constraints, where the sum of all parts equals the total number of unsatisfied constraints.

\begin{figure*}
    \centering
    \includegraphics[width=0.98\linewidth]{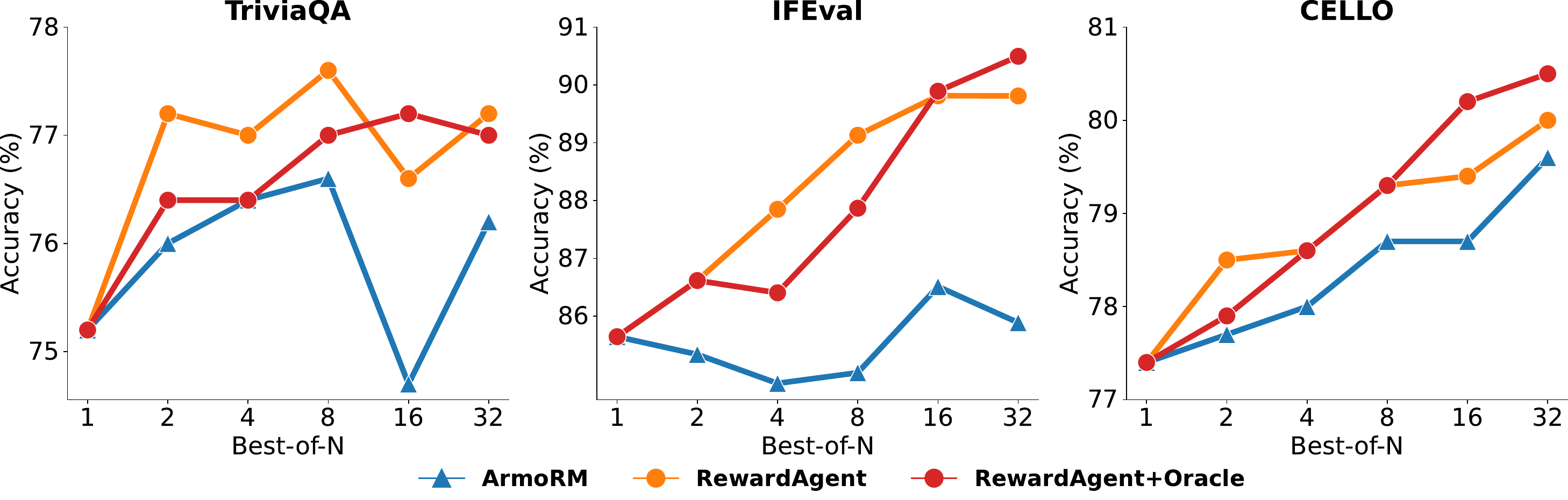}
    \caption{Best-of-n results (\%) on TriviaQA, IFEval, and CELLO using the base reward model ArmoRM and \ourmethod to search. ``+Oracle'' denotes using the oracle setting of \ourmethod as mentioned in \cref{sec:exp_analysis}.}
    \label{fig:gpt4o_best_of_n}
\end{figure*}

\subsection{Evaluation Details}
\label{sec:app_exp_evaluation}

\paragraph{Best-of-N} For the TriviaQA, we sample $500$ instances from the validation split in \texttt{rc.nocontext} version. The model is prompted to generate direct answers, and we report the exact match accuracies. For the IFEval, we report the average accuracy across the strict prompt, strict instruction, loose prompt, and loose instruction settings. For the CELLO, we report the average score based on the official evaluation script. All three tasks are conducted under a zero-shot setting.

\paragraph{DPO Training}
For MT-Bench and CELLO, we employ FastChat\footnote{\url{https://github.com/lm-sys/FastChat/tree/main/fastchat/llm_judge}} and the official evaluation script respectively, to conduct the evaluations and report the average scores.
For the other tasks, we use the \texttt{lm-evaluation-harness}\footnote{\url{https://github.com/EleutherAI/lm-evaluation-harness}} for evaluation. Specifically, we adopt a 5-shot setting for the MMLU and MMLU-Pro tasks, while using a zero-shot setting for TriviaQA and TruthfulQA. Notably, for TruthfulQA, we use the \texttt{truthfulqa\_gen} setting. 

\subsection{More Results on Best-of-N}
\label{sec:app_exp_more_res}
We conduct best-of-n search experiments using \texttt{gpt-4o-2024-11-20} as the policy model, with the results presented in Figure~\ref{fig:gpt4o_best_of_n}. The results demonstrate that \ourmethod significantly improves best-of-n performance compared to the base reward model ArmoRM, even when applied to a more powerful policy model than \ourmethod.

\begin{table*}
    \centering
    \small
    \begin{adjustbox}{max width=1\linewidth}
    {
    \begin{tabular}{p{\linewidth}}
    \toprule
   Given the following instruction, determine whether the following check in needed. \\
    \\
        \text{[Instruction]} \\
        \{instruction\} \\
    \\
        \text{[Checks]} \\
       \{ 
            ``name'': ``constraint check'', 
            ``desp'': ``A `constraint check' is required if the instruction contains any additional constraints or requirements on the output, such as length, keywords, format, number of sections, frequency, order, etc.'', 
            ``identifier'': ``[[A]]'' 
        \}, 
        \{  
            ``name'': ``factuality check'', 
            ``desp'': ``A `factuality check' is required if the generated response to the instruction potentially contains claims about factual information or world knowledge.'', 
            ``identifier'': ``[[B]]'' 
        \} \\
        \\
        If the instruction requires some checks, please output the corresponding identifiers (such as [[A]], [[B]]). \\
        Please do not output other identifiers if the corresponding checkers not needed. \\
    \bottomrule
    \end{tabular}
    }
    \end{adjustbox}
    \caption{Our prompt for the router, where the \{instruction\} part varies based on the input. }
    \label{tab:planner}
\end{table*}

\begin{table*}
    \centering
    \small
    \begin{adjustbox}{max width=1\linewidth}
    {
    \begin{tabular}{p{\linewidth}}
    \toprule
    \textbf{Prompt For Difference Proposal} \\
        \text{[Answers]} \\
        \{formatted\_answers\} \\
        \\
        \text{[Your Task]} \\
        Given the above responses, please identify and summarize one key points of contradiction or inconsistency between the claims. \\
        \\
        \text{[Requirements]} \\
        1. Return a Python list containing only the most significant differences between the two answers. \\
        2. Do not include any additional explanations, only output the list. \\
        3. If there are no inconsistencies, return an empty list. \\
    \midrule
    \textbf{Prompt For Query Generation} \\
    \text{[Original question that caused the inconsistency]} \\
        \{instruction\} \\
\\
        \text{[Inconsistencies]} \\
        \{inconsistencies\} \\
        \\
        \text{[Your Task]} \\
        To resolve the inconsistencies, We need to query search engine. For each contradiction, please generate a corresponding query that can be used to retrieve knowledge to resolve the contradiction.  \\
        \\
        \text{[Requirements]} \\
        1. Each query should be specific and targeted, aiming to verify or disprove the conflicting points.  \\
        2. Provide the queries in a clear and concise manner, returning a Python list of queries corrresponding to the inconsistencies. \\
        3. Do not provide any additional explanations, only output the list. \\
        \midrule
    \textbf{Prompt For Verification} \\
    Evaluate which of the two answers is more factual based on the supporting information. \\
        \text{[Support knowledge sources]}: \\
        \{supports\} \\
        \\
        \text{[Original Answers]}: \\
        \{formatted\_answers\} \\
        \\
        \text{[Remeber]} \\
        For each answer, provide a score between 1 and 10, where 10 represents the highest factual accuracy. Your output should only consist of the following: \\
        Answer A: [[score]] (Wrap the score of A with [[ and ]]) \\
        Answer B: <<score>> (Wrap the score of B with << and >>) \\
        Please also provide a compact explanation. \\
    \bottomrule
    \end{tabular}
    }
    \end{adjustbox}
    \caption{Our prompt for assessing factuality in verification agents, with the \{formatted\_answers\}, \{supports\}, \{inconsistencies\}, \{instruction\} and \{supports\} parts varying based on the input. }
    \label{tab:factuality_agent}
\end{table*}

\begin{table*}
    \centering
    \small
    \begin{adjustbox}{max width=1\linewidth}
    {
    \begin{tabular}{p{\linewidth}}
    \toprule
    \textbf{Prompt For Constraint Parsing} \\
       You are an expert in natural language processing and constraint checking. Your task is to analyze a given instruction and identify which constraints need to be checked. \\
        \\
        The `instruction' contains a specific task query along with several explicitly stated constraints. Based on the instructions, you need to return a list of checker names that should be applied to the constraints. \\
        \\
        Task Example: \\  
        Instruction: Write a 300+ word summary of the Wikipedia page ``https://en.wikipedia.org/wiki/Raymond\_III,\_Count\_of\_Tripol''. Do not use any commas and highlight at least 3 sections that have titles in markdown format, for example, *highlighted section part 1*, *highlighted section part 2*, *highlighted section part 3*.\\
        Response: \\
        NumberOfWordsChecker: 300+ word \\
        HighlightSectionChecker: highlight at least 3 sections that have titles in markdown format\\
        ForbiddenWordsChecker: Do not use any commas \\
        \\
        Task Instruction: \\
        \{instruction\} \\
        \\
        \#\#\# Your task: \\
        - Generate the appropriate checker names with corresponding descriptions from the original instruction description. \\
        - Return the checker names with their descriptions separated by `\textbackslash n'  \\
        - Focus only on the constraints explicitly mentioned in the instruction (e.g., length, format, specific exclusions).  \\
        - Do **not** generate checkers for the task query itself or its quality. \\
        - Do **not** infer or output constraints that are implicitly included in the instruction (e.g., general style or unstated rules). \\
        - Each checker should be responsible for checking only one constraint. \\
    \midrule
    \textbf{Prompt For Code Generation} \\
    You are tasked with implementing a Python function `check\_following' that determines whether a given `response' satisfies a constraint defined by a checker. The function should return `True' if the constraint is satisfied, and `False' otherwise. \\
\\
        \text{[Instruction to check]}: \\
        \{instruction\} \\
\\
        \text{[Specific Checker and Description]}: \\
        \{checker\_name\} \\
\\
        Requirements: \\
        - The function accepts only one parameter: `response' which is a Python string. \\
        - The function must return a boolean value (`True' or `False') based on whether the `response' adheres to the constraint described by the checker. \\
        - The function must not include any I/O operations, such as `input()' or `ArgumentParser'. \\
        - The Python code for each checker should be designed to be generalizable, e.g., using regular expressions or other suitable techniques. \\
        - Only return the exact Python code, with no additional explanations. \\
    \bottomrule
    \end{tabular}
    }
    \end{adjustbox}
    \caption{Our prompt for assessing instruction-following in verification agents, with the \{instruction\} and \{checker\_name\} parts varying based on the input. }
    \label{tab:if_agent}
\end{table*}

\end{document}